\documentclass[runningheads]{llncs}
\usepackage[namelimits]{amsmath} 
\usepackage{amssymb}             
\usepackage{amsfonts}            
\usepackage{mathrsfs}  
\usepackage{graphicx}
\usepackage{booktabs}
\usepackage{xcolor}
\usepackage[misc]{ifsym} 
\usepackage[T1]{fontenc}
\usepackage[utf8]{inputenc}
\usepackage{authblk}
\begin{document}
\title{Medical Report Generation based on Segment-Enhanced Contrastive Representation Learning}

\author{Ruoqing Zhao, \ Xi Wang, \ Hongliang Dai, \ Pan Gao, \ 
Piji Li\thanks{Corresponding author}}

% Nanjing University of Aeronautics and Astronautics \\
% \texttt{\{rqzhao,\ xiwang0102,\ hongldai, \ Pan.Gao, 
% \ Pjli\}@nuaa.edu.cn}\\},  

\authorrunning{R. Zhao et al.}

\institute{ College of Computer Science and Technology, \\Nanjing University of Aeronautics and Astronautics\\MIIT Key Laboratory of Pattern Analysis and Machine Intelligence \\
\email{\{rqzhao,\ xiwang0102,\ hongldai,\ pan.gao,\ pjli\}@nuaa.edu.cn}}

\maketitle              % typeset the header of the contribution
\begin{abstract}
Automated radiology report generation has the potential to improve radiology reporting and alleviate the workload of radiologists. However, the medical report generation task poses unique challenges due to the limited availability of medical data and the presence of data bias. 
% To maximize the utility of available data and alleviate data bias, we propose a Medical images Segment with Contrastive Learning framework (MSCL) that utilizes the Segment Anything Model (SAM) to segment the organs, abnormalities, bones, and others, that could pay more attention to abnormal regions or lesions to get better visual representation. 
%\revisedhl{To maximize the utility of available data and reduce data bias, we propose MSCL (Medical image Segmentation with Contrastive Learning), a framework that utilizes the Segment Anything Model (SAM) to segment organs, abnormalities, bones, etc., and can pay more attention to the meaningful ROIs in the image to get better visual representations.}
To maximize the utility of available data and reduce data bias, we propose MSCL (Medical image Segmentation with Contrastive Learning), a framework that utilizes the Segment Anything Model (SAM) to segment organs, abnormalities, bones, etc., and can pay more attention to the meaningful ROIs in the image to get better visual representations. Then we introduce a supervised contrastive loss that assigns more weight to reports that are semantically similar to the target while training. The design of this loss function aims to mitigate the impact of data bias and encourage the model to capture the essential features of a medical image and generate high-quality reports. Experimental results demonstrate the effectiveness of our proposed model, where we achieve state-of-the-art performance on the IU X-Ray public dataset.

\end{abstract}
\section{Introduction}
% Automated radiology report generation is an innovative approach that aims to generate informative text from radiologic image studies. This technology has the potential to improve radiology reporting and alleviate the workload of radiologists. However, due to a lack of qualified radiologists, many reports may contain indecisive findings, necessitating further tests involving pathology or other advanced imaging methods. Additionally, the manual process of generating full text radiology reports is time-consuming and poses significant challenges. To address these issues, researchers have turned to deep learning models capable of automated report generation, which offer a promising solution to facilitate the efficient and effective production of diagnostic reports.

%\revisedhl{The process of manually writing full text radiology reports is time-consuming and poses significant challenges. Additionally, the reports written by inexperienced radiologists may contain indecisive findings, resulting in the necessity of further tests involving pathology or other advanced imaging methods. The automated radiology report generation task, which aims to generate informative text from radiologic image studies, has the potential to improve radiology reporting and alleviate the workload of radiologists. (I changed the order of sentences based on my understanding. Please check the correctness.)}
The process of manually writing full text radiology reports is time-consuming and poses significant challenges. Additionally, the reports written by inexperienced radiologists may contain indecisive findings, resulting in the necessity of further tests involving pathology or other advanced imaging methods. The automated radiology report generation task, which aims to generate informative text from radiologic image studies, has the potential to improve radiology reporting and alleviate the workload of radiologists.

\begin{figure}
\centering
\includegraphics[width=1\columnwidth]{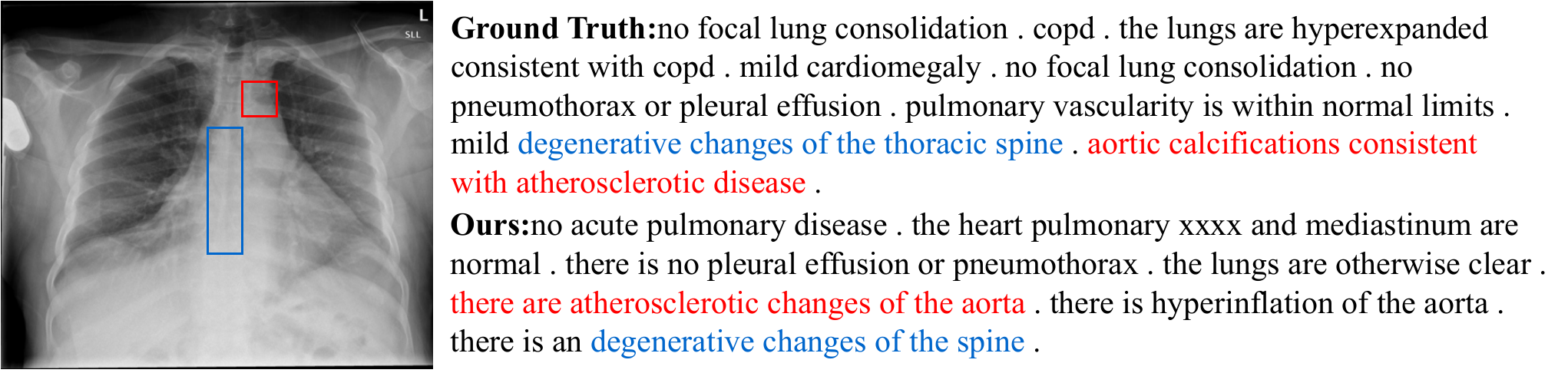}
\caption{One example of Chest X-ray image with the corresponding ground truth and our model generated reports. The abnormal regions and their corresponding descriptions are marked in same colors, showing serious data biases of this task.}\label{fig1}
\vspace{-6mm}
\end{figure}

%\revisedhl{Currently, the mainstream approach to medical report generation is to use a deep-learning-based encoder-decoder architecture~\cite{Wang2018TieNetTE,Jing2017OnTA,Xue2018MultimodalRM}}.
Currently, the mainstream approach to medical report generation is to use a deep-learning-based encoder-decoder architecture~\cite{Wang2018TieNetTE,Jing2017OnTA,Xue2018MultimodalRM}. 
% Many existing deep-learning-based medical report generation methods mainly adopt the encoder-decoder architecture \cite{Wang2018TieNetTE,Jing2017OnTA,Xue2018MultimodalRM}. 
In this architecture, convolutional neural networks (CNNs) are utilized to encode the input medical images, while recurrent neural networks (RNNs), such as long short-term memory (LSTM), or non-recurrent networks (e.g., Transformer~\cite{vaswani2017attention}), are used as decoders to automatically generate medical reports. However, directly applying these approaches to medical images has the following problems: i) Visual data bias is a prevalent issue in medical image analysis~\cite{Shin2016LearningTR}, where the dataset is often skewed towards normal images, leading to a disproportionate representation of abnormal images. Moreover, in abnormal images, normal regions can dominate the image, further exacerbating the problem of bias. ii) Textual data bias is another challenge. As shown in Fig.~\ref{fig1}, radiologists tend to describe all items in an image in their medical reports. This can result in descriptions of normal regions dominating the report, with many identical sentences used to describe the same normal regions. This can also aggravate the problem of visual data bias by reinforcing the over-representation of normal regions in the dataset. Consequently, these two data biases could mislead the model training~\cite{Li2018HybridRR,Xue2018MultimodalRM}. 
%\revisedhl{Problem iii is not clear to me. Are long sentence the cause of incoherence? How is it related to the two data biases?}

% To tackle these challenging issues, we propose an effective but simple Medical images Segment with Constrastive Learning framework (MSCL) for better medical report generation. 
% Specifically, inspired by the success of Segment Anything Model (SAM) \cite{kirillov2023segment}, showing impressive zero-shot inference performance in different domains, we adopt it to segment the organs, abnormalities, bones, and others, that could pay more attention to abnormal regions or lesions to get better visual representations. 
%\revisedhl{To tackle these challenging issues, we propose a simple but effective framework called MSCL (Medical images Segment with Constrastive Learning) for better medical report generation. Specifically, inspired by the impressive zero-shot inference performance of Segment Anything Model (SAM)~\cite{kirillov2023segment}, we adopt it to segment organs, abnormalities, bones, and others. }
To tackle these challenging issues, we propose a simple but effective framework called MSCL (Medical images Segment with Contrastive Learning) for better medical report generation. Specifically, inspired by the impressive zero-shot inference performance of Segment Anything Model (SAM)~\cite{kirillov2023segment}, we adopt it to segment organs, abnormalities, bones, and others. 
Given a medical image, we first use the SAM to perform fine-grained segmentation of medical images, focusing on meaningful ROIs that may contain abnormalities in the image, and then extract the image features from these segmentation. 
%\revisedhl{This allows the model to pay more attention to the regions where diseases may exist rather than other meaningless regions to get better visual representations.}
This allows the model to pay more attention to the regions where diseases may exist rather than other meaningless regions to get better visual representations, facilitating a more targeted and precise analysis. 
Furthermore, to mitigate the text data bias issue, we introduce a supervised contrastive loss during the training process. This loss function encourages the model to distinguish between target reports and erroneous ones, assigning more weight to reports that accurately describe abnormalities. By emphasizing the contrast between different report instances, we alleviate the dominance of normal region descriptions and promote more balanced and informative reports. 
%In addition, contrastive learning can help learn robust representations which help capture the subtlety of visual features required for medical image understanding tasks and generate more diverse and accurate reports.
Experimental results on a public dataset, IU-Xray~\cite{DemnerFushman2015PreparingAC}, confirm the validity and effectiveness of our proposed approach. 
%\revisedhl{(I do not see how the approach addresses challenge ii and iii.)}

Overall, the main contributions of this work are:
\begin{itemize}
\item[$\bullet$] We improve visual representations by segmenting meaningful ROIs of the image via applying the Segment Anything Model to medical report generation.
\item[$\bullet$] We propose an effective objective for training a chest X-ray report generation model with a contrastive term. It effectively contrasts target reports with erroneous ones during the training process to alleviate the data bias. 
%\revisedhl{How contrastive learning helps to alleviate the data bias is not explained previously.}
\item[$\bullet$] We conduct comprehensive experiments to demonstrate the effectiveness of our proposed method, which outperforms existing methods on text generation metrics.
\end{itemize}

\section{Related Work}
\subsection{Medical Report Generation}
The mainstream paradigm of medical report generation is the encoder-decoder architecture. Inspired by image captioning~\cite{Vinyals2015ShowandTell}, the early works of medical report generation use CNN-RNN framework~\cite{Shin2016LearningTR,Wang2018TieNetTE,Li2020RigidFC,Xue2018MultimodalRM}. As mentioned in~\cite{Jing2017OnTA}, RNNs are incapable of generating long sentences and paragraphs. To tackle this issue, some~\cite{Jing2017OnTA,Xue2018MultimodalRM} choose hierarchical RNN architectures to produce high-quality long texts, others~\cite{Chen2020GeneratingRR,Nguyen2021AutomatedGO} turn to use Transformer~\cite{vaswani2017attention} as the text decoder of the model. 
%Reinforcement learning~\cite{DBLP:books/lib/SuttonB98} is also adopted in the medical report generation area.
\cite{Li2018HybridRR} uses a hybrid model with a retrieval module and a generation module to generate normal and abnormal sentences respectively. 
The use of prior medical knowledge such as knowledge graph is also exploited in~\cite{zhang2020whenRadiology}, which extracts disease keywords from the reports as nodes in the chest abnormality
graph, therefore facilitates the model's learning of each disease.

\subsection{Contrastive Learning}
Contrastive learning has been widely applied in many fields of machine learning. By contrasting between positive and negative pairs, models can learn a better image representation~\cite{10.1145/3543507.3583285}. Inspired by previous works of contrastive learning in medical images~\cite{moco,SimCLR}, many works have been done to improve the performance of medical report generation in different ways. Chen et al.~\cite{chen2023representative} finds out that many current models using image decoders pretrained with datasets of different domains, which fail to learn the specific image representations in the medical domain. They then use contrastive study to optimize the image representations. 

\subsection{Segment Anything Model}
Segment Anything Model (SAM)~\cite{kirillov2023segment} is Transformer-based model for image segmentation that raises the promptable segmentation task, which is aimed to return a segmentation mask given prompt including spatial or text information. SAM consists of a Vision Transformer based image encoder to extract image embeddings, a prompt encoder to generate prompt embeddings from various kinds of prompts and a mask decoder to output the valid masks and their corresponding confidence scores. Recently there are works on improving the performance of SAM in the medical domain~\cite{MedSAM} which builds a large medical image dataset and proposes an approach for fine-tuning the model to adapt to the medical domain.

\section{Method}
\begin{figure}
\vspace{-4mm}
\centering
\includegraphics[width=1\columnwidth]{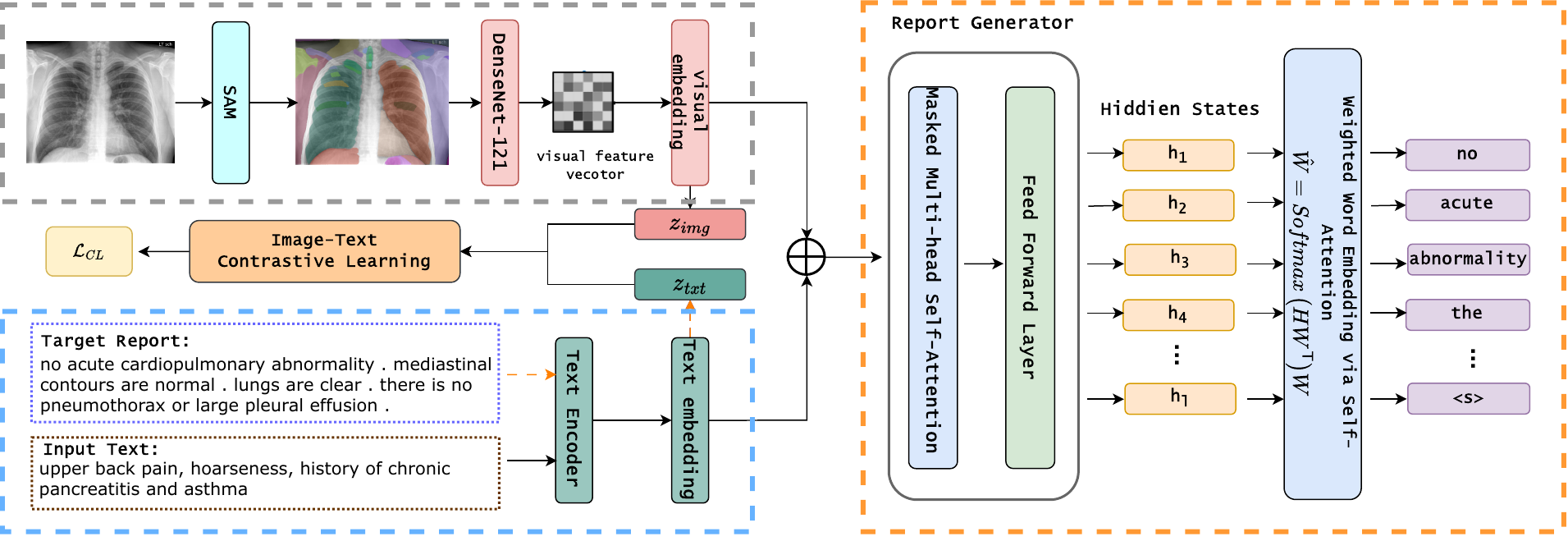}
\caption{ The overall architecture of our proposed model, including three main modules: Visual Extractor, Text Encoder and Report Generator. They are shown in grey dash boxes, blue dash boxes and orange dash boxes, respectively. In addition, Segment medical images with SAM Module is included in Visual Extractor Module and Image-Text Contrastive Learning is adopted for training MSCL. }\label{fig2}
\vspace{-6mm}
\end{figure}

In this section, we present the details of the proposed method. The overall structure of MSCL is illustrated in Fig.~\ref{fig2}, which contains three basic modules and two proposed modules. 
% We first describe the overview of Medical images Segment with Constrastive Learning (MSCL), and then introduce Segment medical images with SAM Module and Image-Text Contrastive Learning Module, respectively.
%\revisedhl{We first describe the overview of Medical images Segment with Constrastive Learning (MSCL), and then introduce the SAM Medical Image Segmentation module and the Image-Text Contrastive Learning module, respectively.}
We first describe the background of Medical images Segment with Contrastive Learning (MSCL), and then introduce the SAM Medical Image Segmentation module and the Image-Text Contrastive Learning module, respectively.

\subsection{Background}
In this work, we leverage the Transformer framework proposed in~\cite{Nguyen2021AutomatedGO}, an end-to-end approach, as our backbone model to generate fluent and robust report. The overall description of the three modules and the training objective is detailed below. 

\noindent\textbf{Visual Extractor} Since each medical study consists of $m$ chest X-ray images $\left \{ X_i \right \} ^m _{i=1}$, its visual latent features $\left \{ x_i \right \} ^m _{i=1} \in \mathbb{R}^c $ are extracted by a shared DenseNet-121~\cite{8099726} image encoder, where $c$ is the number of features. Then, the global visual feature representation $x \in \mathbb{R}^c $ can be obtained by max-pooling across the set of $m$ visual latent features 
$\left \{ x_i \right \} ^m _{i=1}$, 
%\revisedhl{$\left \{ x_i \right \} ^m _{i=1}$,}
as proposed in~\cite{7410471} . The global visual feature representation $\left \{ x_i \right \} ^m _{i=1} \in \mathbb{R}^c $ are subsequently decoupled into low-dimensional disease representations and regarded as the \emph{visual embedding} $\mathbf{D}_{img}  \in \mathbb{R}^{n \times d } $, where each row is a vector $\mathbf{\phi}_j\left ( x \right )  \in \mathbb{R}^d $, $ j = 1, \ldots , n$ defined as follows:
\begin{equation}
\phi _j(x) = \mathbf{A}^\mathsf{T} _j x + b_j
\end{equation}
where $\mathbf{A}_j\in \mathbb{R}^{c \times d}$ and $b_j\in \mathbb{R}^{d}$ are learnable parameters of the $j$-th disease representation. $n$ is the number of disease representations, and $d$ is the embedding dimension.

\noindent\textbf{Text Encoder} In our model, we use the Transformer encoder ~\cite{vaswani2017attention} as our text feature extractor. Denote its output
hidden states as $ \mathbf{H} =\left \{ h_1, h_2,\ldots, h_l \right \} $, where $h_i\in \mathbb{R}^{d}$ is the attended features of the $i$-th word to other words in the text,
\begin{equation}
h_i =f_e\left ( w_i\mid w_1,w_2,\cdots ,w_l \right ) 
\end{equation}
where $f_e$ refers to the encoder, $w_i\in \mathbb{R}^{d}$ stands for the $i$-th word in the text and $l$ is the length of the text.
% The entire report T is then summarized by $ Q =\left \{ q_1, q_2,\ldots, q_n \right \} $ where $q\in \mathbb{R}^{d}$ and then learned via the attention process, representing n disease topics (e.g., pneumonia or atelectasis) mentioned in CheXpert~\cite{Irvin2019CheXpertAL}. Then the \textit{text embedding} $D_{txt} \in \mathbb{R}^{n \times d } $ can be written as follows:
%\revisedhl{Following CheXpert~\cite{Irvin2019CheXpertAL}, the entire report T is then summarized based on $n$ disease topics (e.g., pneumonia or atelectasis) represented with $ Q =\left \{ q_1, q_2,\ldots, q_n \right \} $, where $q\in \mathbb{R}^{d}$.  The \textit{text embedding} $D_{txt} \in \mathbb{R}^{n \times d } $ is computed via attention as follows:}
Following CheXpert~\cite{Irvin2019CheXpertAL}, the entire report T is then summarized based on $n$ disease topics (e.g., pneumonia or atelectasis) represented with $ \mathbf{Q} =\left \{ q_1, q_2,\ldots, q_n \right \} $, where $q\in \mathbb{R}^{d}$.  The \textit{text embedding} $\mathbf{D}_{txt} \in \mathbb{R}^{n \times d } $ is computed via attention as follows:
\begin{equation}
\mathbf{D}_{txt} = Softmax\left (\mathbf{Q}  \mathbf{H}^\mathsf{T} \right ) \mathbf{H}
\end{equation}
where $\mathbf{Q}\in \mathbb{R}^{n \times d}$ and $\mathbf{H}\in \mathbb{R}^{l \times d}$ is formed by $ \left \{ h_1, h_2,\ldots, h_l \right \} $ from Eq. (2). The term $Softmax\left ( \mathbf{Q}  \mathbf{H}^\mathsf{T} \right )$ is used to calculate the word attention heat-map for $n$ disease topics in the report.
%Therefore the weighted sum of these words by Eq. (3) gives the feature that summarizes the queried disease in the report.

\noindent\textbf{Report Generator} 
Then we fuse the \textit{text embedding} $\mathbf{D}_{txt}$ and \textit{visual embedding} $\mathbf{D}_{img}$ to get a more comprehensive and representative feature vector $\mathbf{D}_{it} \in \mathbb{R}^{n \times d } $,
\begin{equation}
\mathbf{D}_{it} = LayerNorm\left (\mathbf{D}_{img}+ \mathbf{D}_{txt}\right )
\end{equation}
Followed by~\cite{Nguyen2021AutomatedGO}, in order to get explicit and precise disease descriptions for subsequent generation, 
%we use $D_{states}$ and $D_{topics}$ to represent the disease states and disease-related topics, respectively. 
we let $\mathbf{S}\in \mathbb{R}^{k \times d}$ represent the state embedding where $k$ refers to the number of states, such as \textit{positive, negative, uncertain, or unmentioned}, and $\mathbf{S}\in \mathbb{R}^{k \times d}$ is randomly initialized, then learned via the classification of $\mathbf{D}_{it}$. The confidence of classifying each disease into one of the $k$ disease states is
\begin{equation}
p = Softmax\left ( \mathbf{D}_{it} \mathbf{S}^\mathsf{T} \right ) 
\end{equation}
Then the classification loss is computed as:
\begin{equation}
\mathcal{L}_\mathrm{C}=- \frac{1}{n} \sum_{i=1}^{n} \sum_{j=1}^{k}y_{ij}log\left ( p_{ij} \right )  \end{equation}
where $y_{ij}\in \left \{ 0,1 \right \} $ and $p_{ij}\in \left ( 0,1 \right ) $ are the $j$-th ground-truth and predicted values for the disease $i$-th, respectively. 
% Then $D_{states}$ are computed as:
% \begin{equation}
% D_{states}= \left \{ \begin{array}{ll}     yS,                    & \text{ if training phase}\\      pS,                                 & \text{otherwise} \end{array} \right.  
% \end{equation}
% here $y_{ij}$ the one-hot ground-truth labels about the disease-related topics and $p_{ij}$ is given by Eq. (5). Like the disease queries $Q$, $D_{topics} \in \mathbb{R}^{n \times d } $ represents diseases or topics to be generated which is learned in training through the medical report generation pipeline. Then the final disease embedding is formed as:
% \begin{equation}
% D_{final}= D_{ig}+D_{states}+D_{topics}
% \end{equation}

% Our report generator is derived from the transformer encoder. 
%\revisedhl{Our report generator employs a transformer based decoder.} 
Our report generator employs a Transformer-based decoder. 
The network is formed by sandwiching  stacking a masked multi-head self-attention component and a feed-forward layer being on top of each other for $N$ times. The hidden state for each word position $h_i\in \mathbb{R}^{d}$ in the medical report is then computed based on previous words and disease embedding, as $\mathbf{D}_{it}= \left \{ d_i \right \} ^n _{i=1}$, 
\begin{equation}
% h_i =f_e\left ( w_1\mid w_1,w_2,\cdots ,w_{i-1}, d_1,d_2,\cdots ,d_n\right ) 
h_i =f_d\left (w_1,w_2,\cdots ,w_{i-1}, d_1,d_2,\cdots ,d_n\right )
\end{equation}
%\revisedhl{(This equation seems problematic, $h_i =f_d\left (w_1,w_2,\cdots ,w_{i-1}, d_1,d_2,\cdots ,d_n\right ) $ ?)}
Let $p_{word,ij}$ denotes the confidence of selecting the $j$-th word in the vocabulary $\mathbf{W}$ for the $i$-th position in the generated medical report,
\begin{equation}
p_{word} = Softmax\left ( \mathbf{H} \mathbf{W}^\mathsf{T} \right ) 
\end{equation}
where the hidden state $\mathbf{H}=\left \{ h_i \right \} ^l_{i=1}$ and $\mathbf{W} \in \mathbb{R}^{v \times d } $ is the vocabulary embedding, $v$ is the vocabulary size. Then  generator loss is defined as a cross entropy of the ground-truth words $y_{word}$ and $p_{word}$,
\begin{equation}
\mathcal{L}_{CE}=- \frac{1}{l} \sum_{i=1}^{l} \sum_{j=1}^{v}y_{word,ij}log\left ( p_{word,ij} \right )
\end{equation}
Finally the weighted word embedding is: 
\begin{equation}
% Y = p_{word}W 
\hat{\mathbf{W}} = p_{word}\mathbf{W} 
\end{equation}
%\revisedhl{($Y$ is not a word)}

\subsection{Segment medical images with SAM}
The Segment Anything Model (SAM) is a Transformer-based architecture, which demonstrates its excellent capabilities in image segmentation tasks. Specifically, SAM employs a vision Transformer-based image encoder to extract image features. These features are then fused with prompt encoders to integrate user interactions and generate a comprehensive representation. The resulting image embedding is then fed into a mask decoder to produce segmentation outcomes and associated confidence scores. 

\begin{figure}
\centering
\includegraphics[width=1\columnwidth]{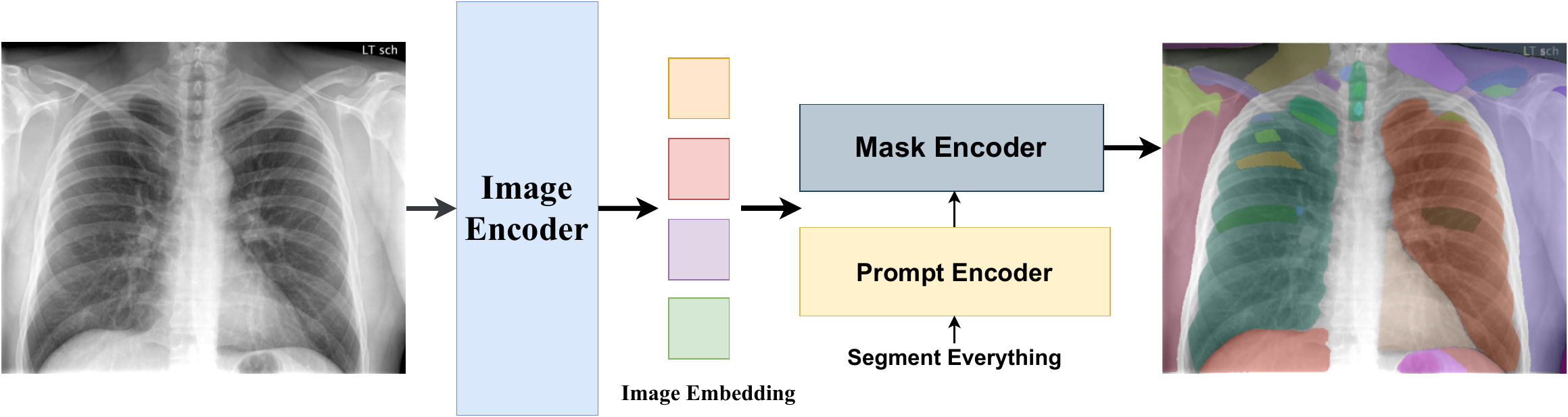}
\caption{Overview of the architecture of Segment Anything Model (SAM).}\label{fig3}
\vspace{-6mm}
\end{figure}

SAM supports three main segmentation modes: segmenting everything in a fully automatic way, bounding box mode, and point mode. Although the bounding box mode and point mode could generate more reasonable segmentation results, since the chest X-ray datasets do not have any annotation and it's a automated process with no human interaction, we adopt the segment-everything mode as our segmentor.

In the everything mode, SAM produces segmentation masks for all the potential objects in the whole image without any manual priors. as shown in Fig.~\ref{fig3}. Given an image $Img$, SAM initially generates a grid of point prompts that covers the entire image. The prompt encoder will produce the point embedding and integrate it with the image embedding using the evenly sampled grid points. Then, using the combination as input, the mask decoder will produce a number of candidate masks for the entire image. The removal of duplicate and poor-quality masks is then accomplished by the application of a filtering method that makes use of confidence scores, stability evaluation based on threshold jitter, and non-maximal suppression approaches. 
\begin{equation}
Img_{pro} = f_{S}\left ( Img\right )
\end{equation}
where $f_{S}$ refers to the SAM segment process. We subsequently apply the processed image $Img_{pro}$ to the visual extractor to extract features.

\subsection{Image-Text Contrastive Learning}
To regularize the training process, we adopt a image-text contrastive loss. We first project the hidden representations of the image and the target sequence into a latent space:
\begin{equation}
\mathit{z_{img}} = \phi _{img}\left ( \tilde{\mathbf{H}_{img}}  \right ),
\mathit{z_{txt}} = \phi _{txt}\left ( \tilde{\mathbf{H}_{txt}}  \right ) 
\end{equation}
where $\tilde{\mathbf{H}_{img}} $ and $\tilde{\mathbf{H}_{txt}} $ are the average pooling of the hidden states $\mathbf{H}_{img}$ and $\mathbf{H}_{txt}$ aforementioned, $\phi _{img}$ and $\phi _{txt}$ are two fully connected layers with ReLU activation~\cite{Nair2010RectifiedLU}. For a batch of paired images and reports, the positive or negative report is based on aforementioned disease topic $\mathbf{D}_{txt}$ in Eq. (3). If they are the same topic with the target report, they are positive samples, else others are negative samples. We then maximize the similarity between the pair of source image and target sequence, while minimizing the similarity between the negative pairs as follows:
\begin{equation}
\mathcal{L}_{CL}=- \sum_{i=1}^{N}log\frac{\exp\left ( s_{i,i} \right )  }{\sum_{l_i\neq l_j}^{}\exp\left ( s_{i,j} \right )+  \theta\sum_{l_i=  l_j}^{}\exp\left ( s_{i,j} \right )  } 
\end{equation}
where $s_{i,j}=\cos (z_{img}^{(i)},z_{txt}^{(j)})/\tau $, $cos$ is the cosine similarity between two vectors, $\tau $ is the temperature parameter, and $\theta $ is a hyperparameter that weighs the importance of negative samples that are semantically close to the target sequence, with the same label $l_i = l_j$ in Eq. (13), here the label refers aforementioned $q\in \mathbb{R}^{d}$ representing $n$ disease topics (e.g., pneumonia or atelectasis). By introducing $\theta $, we focus more on abnormal regions, thereby avoiding generating excessive descriptions for the normal regions in chest X-rays.

Overall, the model is optimized with a mixture of cross-entropy loss and contrastive loss:
\begin{equation}
\mathcal{L}_{total}=\lambda \left ( \mathcal{L}_{c} +\mathcal{L}_{CE}\right ) +\left ( 1-\lambda  \right )\mathcal{L}_{CL} 
\end{equation}
where $\lambda $ is a hyperparameter that weighs the losses.

\section{Experiments}
\subsection{Experimental Settings}
\textbf{Datasets} We conduct experiments on a widely-used radiology reporting benchmarks, IU-Xray~\cite{DemnerFushman2015PreparingAC}, which is collected by the Indiana University hospital network. It contains 7,470 chest images and 3,955 corresponding reports. Either frontal or frontal and lateral view images are associated with each report. Each study typically
consists of \textit{impression}, \textit{findings}, \textit{comparison}, and \textit{indication} sections, we utilize both the multi-view chest X-ray images (frontal and lateral) and the indication section as our inputs. For generating medical reports, we follow the existing literature~\cite{Jing2017OnTA,Srinivasan2020Hierachial} by concatenating the \textit{impression} and the \textit{findings} sections as the
target output. Moreover, we apply the same setting as R2Gen~\cite{Chen2020GeneratingRR} that partition the dataset into train/validation/test set by 7:1:2.

\noindent\textbf{Baseline and Evaluation Metrics} We compare our MSCL with eight state-of-the-art image captioning and medical report generation models as baselines, including ST~\cite{Vinyals2015ShowandTell}, HRGP~\cite{Li2018HybridRR}, TieNet~\cite{Wang2018TieNetTE}, R2Gen~\cite{Chen2020GeneratingRR}, CoAtt~\cite{Jing2017OnTA}, HRG-Transformer~\cite{Srinivasan2020Hierachial} and CMCL~\cite{Liu2022CompetencebasedMC}. We adopt the widely used NLG metrics, including BLEU~\cite{Papineni2002BleuAM}, ROUGE-L~\cite{Lin2004ROUGEAP} and METEOR~\cite{Banerjee2005METEORAA}. Specifically, ROUGE-L is proposed for automatic evaluation of the extracted text summarization. METEOR and BLEU are originally designed for machine translation evaluation.

\noindent\textbf{Implementation Details} We use the DenseNet-121~\cite{8099726} pre-trained on ImageNet to extract visual features of images and the pre-trained ViT-Base model as encoders to run the everything modes. The initial learning rate is set as 3e-4 and the optimizer is AdamW~\cite{loshchilov2017decoupled} with a weight decay of 0.02. We set the weighting parameters $\lambda $ and $\theta $ to 0.8 and 2 respectively. Moreover, we project all encoded vectors by a linear transformation layer into the dimension of $d = 256$.

\begin{table}
\centering
\caption{The performances of our proposed DCL compared with other state-of-the-art systems on IU-Xray dataset. The best results in each column are highlighted in bold.}\label{tab1}
\begin{tabular}{ccccccc} 
\toprule  
    Methods & BLEU-1 & BLEU-2 & BLEU-3 & BLEU-4 & ROUGE-L & METEOR \\
  \midrule 
  ST~\cite{Vinyals2015ShowandTell} & 0.316 & 0.211  & 0.140 & 0.095  & 0.267 & 0.159 \\
  HRGP~\cite{Li2018HybridRR} & 0.438 & 0.298 & 0.208 & 0.151 & 0.322 & - \\
  TieNet~\cite{Wang2018TieNetTE} & 0.330 & 0.194 & 0.124 & 0.081 & 0.311 & - \\
  R2Gen~\cite{Chen2020GeneratingRR} & 0.470 & 0.304 & 0.219 & 0.165 & 0.371 & 0.187 \\ 
  CoAtt~\cite{Jing2017OnTA} & 0.455 & 0.288 & 0.205 & 0.154 & 0.369 & - \\
  HRG-Transformer~\cite{Srinivasan2020Hierachial} & 0.473 & 0.305 & 0.217 & 0.162 & 0.378 & 0.186 \\
  CMCL~\cite{Liu2022CompetencebasedMC} & 0.464 & 0.301 & 0.212 & 0.158 & - & - \\
  Ours & \textbf{0.485} & \textbf{0.355} & \textbf{0.275} & \textbf{0.221} & \textbf{0.433} & \textbf{0.210} \\
  \bottomrule
  \end{tabular}
\end{table}

\subsection{Main Results}
Table~\ref{tab1} shows experimental results of our proposed MSCL and eight baselines on six natural language generation metrics. As is shown, our MSCL achieves the state-of-the-art  performance on all metrics. The BLEU-1 to BLEU-4 metrics analyze how many continuous sequences of words appear in the predicted reports. In our results, they are significantly improved. Especially, the BLEU-4 is 5.6\% higher than R2Gen. One plausible explanation for the superior performance of our method is that it mitigates visual and textual bias, while placing greater emphasis on the longer phrases used to describe diseases. Additionally, we utilized the ROUGE-L metric to evaluate the fluency and adequacy of our generated reports. Our ROUGE-L score was found to be 5.5\% higher than that of the previous state-of-the-art method, indicating that our approach can produce more accurate reports, rather than repeating frequent sentences. Furthermore, we used the METEOR metric to assess the degree of synonym transformation between our predicted reports and the ground truth. The results of the METEOR evaluation further demonstrate the effectiveness of our framework.
\begin{table}
\centering 
\caption{Ablation studies on the test set of IU X-ray dataset. ``w/o'' is the abbreviation of without. “MSCL” refers to the full model. “SV” stand for using single view. “CL” refers to adopting Image-Text Contrastive Learning and “SAM” means using SAM to segment image. Best performances are highlighted in bold.}\label{tab2}
\begin{tabular}{ccccccc} 
\toprule  
    Methods & BLEU-1 & BLEU-2 & BLEU-3 & BLEU-4 & ROUGE-L & METEOR \\
  \midrule 
 MSCL(SV) & 0.444 & 0.324 & 0.251 & 0.199 & 0.432 & 0.199 \\
 MSCL(w/o CL) & 0.466 & 0.347  & 0.276 & 0.223  & \textbf{0.437} & 0.208 \\
 MSCL(w/o SAM) & 0.474 & 0.336  & 0.254 & 0.199  & 0.417 & 0.201 \\
 MSCL & \textbf{0.485} & \textbf{0.355} & \textbf{0.275} & \textbf{0.221} & 0.433 & \textbf{0.210} \\
  \bottomrule
  \end{tabular}
\end{table}
\subsection{Ablation Study}
To further verify the effectiveness of each component in our proposed method, we conduct ablation studies on the IU-Xray dataset. As shown in Table\ref{tab2}, when choosing a single X-ray image as the sole input, the performance drops dramatically compared to using multiple X-ray images as input. This suggests that using images from multiple views as input can provide more information about the disease. When removing the Image-Text Contrastive Learning module, the performance drops on almost all metrics except ROUGE-L. This implies that Image-Text Contrastive Learning helps learn robust representations that capture the essence of a medical image, better focus on the abnormalities. If we remove the SAM module used for segmenting medical images, we observe that each score has been lowered. This observation indicates that segmenting medical images with SAM helps to focus on salient lesions and proves the importance of visual representation qualities, since biased data in chest X-ray generation datasets can severely compromise representation capabilities. Consequently, the results of the ablative experiment verify the effectiveness of our proposed components.

\begin{figure}[!t]
\centering
\includegraphics[width=1\columnwidth]{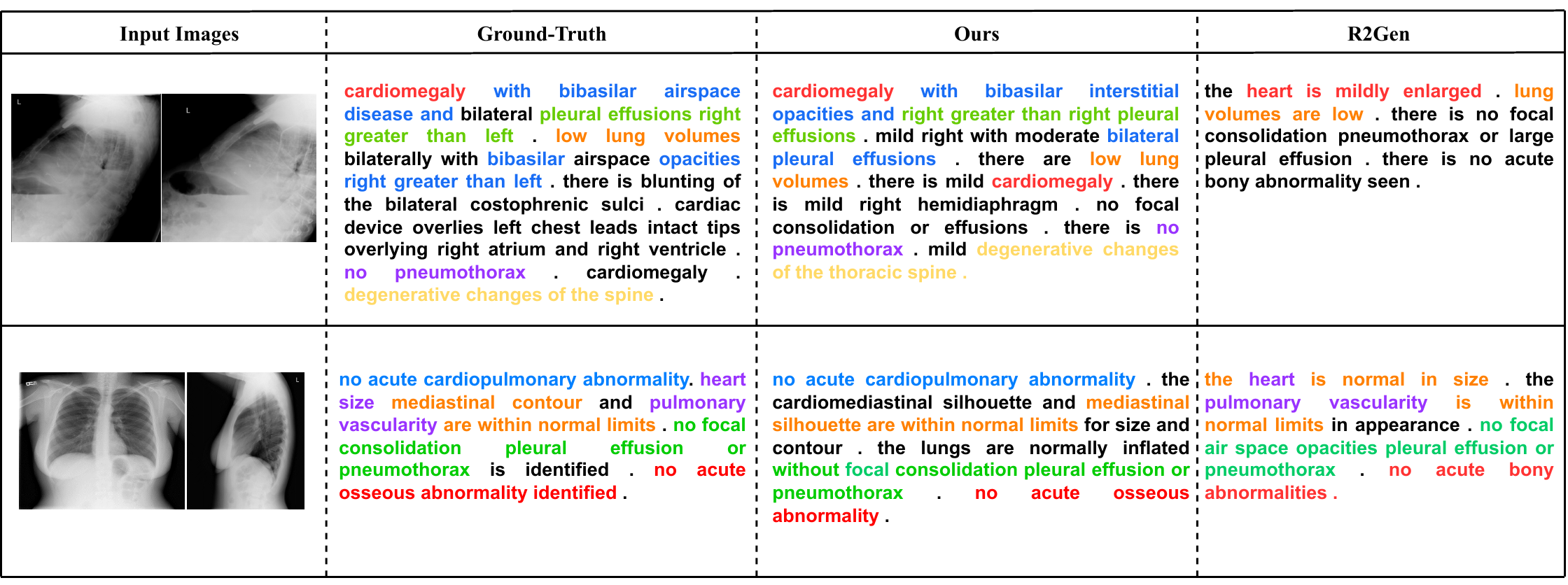}
\caption{ Illustrations of reports from ground-truth, ours and R2Gen~\cite{Chen2020GeneratingRR} for two X-ray chest images. To better distinguish the content in the reports, different colors highlight different medical terms.}\label{fig4}
\vspace{-6mm}
\end{figure}

\subsection{Case Study}
To further investigate the effectiveness of our model, we perform qualitative analysis on some cases with their ground-truth, generated reports from ours and R2Gen~\cite{Chen2020GeneratingRR}. Figure~\ref{fig4} shows two examples of the generated reports in the test set, where different colors on the texts indicate different medical terms. It is observed in these cases that our model generates descriptions that closely align with those written by radiologists in terms of content flow. Specifically, the patterns in the generated reports follow a structured approach, beginning with the reporting of abnormal findings (such as ``cardiopulmonary abnormality" and ``lung volumes"), and concluding with potential diseases (such as "hypoinflation"). Furthermore, when the patient has multiple disease symptoms, as in the first example, we have found that MSCL covers almost all of the necessary medical terms and abnormalities in the ground-truth reports, which proves that the reports generated from our model are comprehensive and accurate compared to R2Gen, which describes more about normal symptoms.
% Please try to avoid rasterized images for line-art diagrams and
% schemas. Whenever possible, use vector graphics instead (see
% Fig.~\ref{fig1}).

\section{Conclusion}
In this paper, we propose an effective but simple Medical images Segment with Contrastive Learning framework (MSCL) to alleviate the data bias by efficiently utilizing the limited medical data for medical report
generation. To this end, we first utilize Segment Anything Model (SAM) to segment medical images, which allows us to pay more attention to the meaningful ROIs in the image to get better visual representations. Then contrastive
learning is employed to expose the model to semantically-close negative samples which improves generation performance. Experimental results demonstrate the effectiveness of our model in generating accurate
and meaningful reports.

\section*{Acknowledgements}
This research is supported by the National Key Research and Development Program of China(No.2021ZD0113203), the National Natural Science Foundation of China (No.62106105), the CCF-Tencent Open Research Fund (No.RAGR20220122), the CCF-Zhipu AI Large Model Fund (No.CCF-Zhipu202315), the Scientific Research Starting Foundation of Nanjing University of Aeronautics and Astronautics (No.YQR21022), and the High Performance Computing Platform of Nanjing University of Aeronautics and Astronautics.

% \begin{figure}
% \includegraphics[width=\textwidth]{fig1.eps}
% \caption{A figure caption is always placed below the illustration.
% Please note that short captions are centered, while long ones are
% justified by the macro package automatically.} \label{fig1}
% \end{figure}

% \begin{theorem}
% This is a sample theorem. The run-in heading is set in bold, while
% the following text appears in italics. Definitions, le  mmas,
% propositions, and corollaries are styled the same way.
% \end{theorem}
%
% the environments 'definition', 'lemma', 'proposition', 'corollary',
% 'remark', and 'example' are defined in the LLNCS documentclass as well.
%
% \begin{proof}
% Proofs, examples, and remarks have the initial word in italics,
% while the following text appears in normal font.
% \end{proof}
% For citations of references, we prefer the use of square brackets
% and consecutive numbers. Citations using labels or the author/year
% convention are also acceptable. The following bibliography provides
% a sample reference list with entries for journal
% articles~\cite{ref_article1}, an LNCS chapter~\cite{ref_lncs1}, a
% book~\cite{ref_book1}, proceedings without editors~\cite{ref_proc1},
% and a homepage~\cite{ref_url1}. Multiple citations are grouped
% \cite{ref_article1,ref_lncs1,ref_book1},
% \cite{ref_article1,ref_book1,ref_proc1,ref_url1}.
\bibliographystyle{splncs04}
\bibliography{custom}

\end{document}